\documentclass[runningheads]{llncs}
\usepackage{graphicx}
\usepackage{comment}
\usepackage{amsmath,amssymb} 
\usepackage{color}
\usepackage{xcolor}
\usepackage{cite}
\usepackage{comment}
\usepackage[misc]{ifsym}
\usepackage{hyperref}
\hypersetup{
    colorlinks=true,
    linkcolor=blue,
    filecolor=magenta,      
    urlcolor=blue,
}

\usepackage{float}
\usepackage{bbm}
\usepackage{bm}

\definecolor{revision}{rgb}{1.0, 0.49. 0.0}

\begin{document}

\pagestyle{headings}
\mainmatter
\def\ECCVSubNumber{612}  

\title{Synthesize then Compare: Detecting Failures and Anomalies for Semantic Segmentation}

\titlerunning{Synthesize then Compare}
%
\author{Yingda Xia\textsuperscript{*}, Yi Zhang\textsuperscript{*}, Fengze Liu, Wei Shen$^{(\textrm{\Letter})}$, Alan L. Yuille}

\index{Xia, Yingda}
\index{Zhang, Yi}
\index{Liu, Fengze}
\index{Shen, Wei}
\index{Yuille, Alan}

\authorrunning{Y. Xia \textit{et al.}}
%
\institute{Johns Hopkins University}
\maketitle
\let\thefootnote\relax\footnote{\text{*} The first two authors equally contributed to the work. $^{\textrm{\Letter}}$ Corresponding Author.\\
Code available at \url{https://github.com/YingdaXia/SynthCP}}
\begin{abstract}
The ability to detect failures and anomalies are fundamental requirements for building reliable systems for computer vision applications, especially safety-critical applications of semantic segmentation, such as autonomous driving and medical image analysis. In this paper, we systematically study failure and anomaly detection for semantic segmentation and propose a unified framework, consisting of two modules, to address these two related problems. The first module is an image synthesis module, which generates a synthesized image from a segmentation layout map, and the second is a comparison module, which computes the difference between the synthesized image and the input image. We validate our framework on three challenging datasets and improve the state-of-the-arts by large margins, \emph{i.e.}, 6\% AUPR-Error on Cityscapes, 7\% Pearson correlation on pancreatic tumor segmentation in MSD and 20\% AUPR on StreetHazards anomaly segmentation.

\keywords{failure detection, anomaly segmentation, semantic segmentation}
\end{abstract}

\section{Introduction}
Deep neural networks~\cite{alexnet, vgg, googlenet, he2016resnet, densenet} have achieved great success in various computer vision tasks. However, when they come to real world applications, such as autonomous driving~\cite{driving}, medical diagnoses~\cite{medicalsafety} and nuclear power plant monitoring~\cite{nuclear}, the safety issue~\cite{aisafety} raises tremendous concerns particularly in conditions where failure cases have severe consequences. As a result, it is of enormous value that a machine learning system is capable of detecting the failures, \emph{i.e.}, wrong predictions, as well as identifying the anomalies, \emph{i.e.}, out-of-distribution (OOD) cases, that may cause these failures.

Previous works on failure detection~\cite{hendrycks2016baseline,trustscore,tcp} and anomaly (OOD) detection~\cite{ood1,ood2,ood3,ood4,ood5} mainly focus on classifying small images. Although failure detection and anomaly detection for semantic segmentation have received little attention in the literature so far, they are more closely related to safety-critical applications, \emph{e.g.}, autonomous driving and medical image analysis. The objective of failure detection for semantic segmentation is not only to determine whether there are failures in a segmentation result, but also to locate where the failures are. Anomaly detection for semantic segmentation, \emph{a.k.a} anomaly segmentation, is related to failure detection, and its objective is to segment anomalous objects or regions in a given image.  

\begin{figure}[!t]
\begin{center}
    \includegraphics[width=\textwidth]{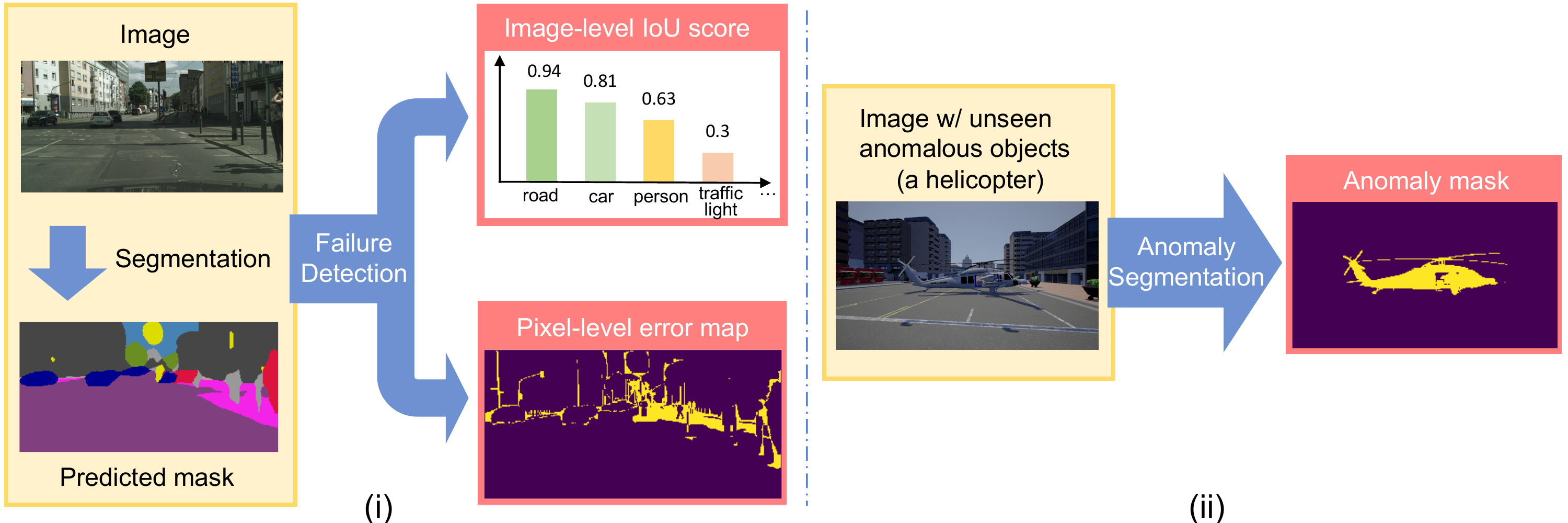}
\end{center}
\caption{
   We aim at addressing two tasks: (i) failure detection, \emph{i.e.}, image-level per-class IoU prediction (top left) and pixel-level error map prediction (bottom left) (ii) anomaly segmentation i.e. segmenting anomalous objects (right middle).
}
\label{Fig1}
\end{figure}

In this paper, our goal is to build a reliable alarm system to address failure detection for semantic segmentation (Fig.~\ref{Fig1}(i)) and anomaly segmentation (Fig.~\ref{Fig1}(ii)). Unlike image classification outputs only a single image label, semantic segmentation outputs a structured semantic layout. Thus, this requires that the system should be able to provide more detailed analysis than those for image classification, \emph{i.e.}, pixel-level error/confidence maps. Some previous works~\cite{Unc17, tcp, caos} directly applied the failure/anomaly detection strategies for image classification pixel by pixel to estimate a pixel-level error map, but they lack the consideration of the structured semantic layout of a segmentation result. 

We propose a unified framework to address failure detection and anomaly detection for semantic segmentation. This framework consists of two components: an image synthesis module, which synthesizes an image from a segmentation result to reconstruct its input image, \emph{i.e.}, a reverse procedure of semantic segmentation, and a comparison module which computes the difference between the reconstructed image and the input image. Our framework is motivated by the fact that the quality of semantic image synthesis~\cite{pix2pix,pix2pixhd,spade} can be evaluated by the performance of segmentation network. Presumably the converse is also true, the better is the segmentation result, the closer a synthesized image generated from the segmentation result is to the input image. If a failure occurs during segmentation, for example, if a person is mis-segmented as a pole, the synthesized image generated from the segmentation result does not look like a person and an obvious difference between the synthesized image and the input image should occur. Similarly, when an anomalous (OOD) object occurs in a test image, it would be classified as any possible in-distribution objects in a segmentation result, and then appear as in-distribution objects in the synthesized image generated from the segmentation result. Consequently, the anomalous object can be identified by finding the differences between the test image and the synthesized image. We refer to our framework as SynthCP, for ``synthesize then compare''.

We model this synthesis procedure by a semantic-to-image conditional GAN (cGAN)~\cite{spade}, 
which is capable of modeling the mapping from the segmentation layout space to the image space. This cGAN is trained on label-image pairs. Given the segmentation result of an input image obtained by an semantic segmentation model, we apply the trained cGAN to the segmentation result to generate a reconstructed image. Then, the reconstructed image and the input image are fed into the comparison module to identify the failures/anomalies. The comparison module is designed task-specifically: For failure detection, the comparison module is modeled by a Siamese network, outputting both image-level confidences and pixel-level confidences; For anomaly segmentation, the comparison module is realized by computing the distance defined on the intermediate features extracted by the semantic segmentation model. 

We validate SynthCP on the Cityscapes street scene dataset, a pancreatic tumor segmentation dataset in the Medical Segmentation Decathlon (MSD) challenge and the StreetHazards dataset, and show its superiority to other failure detection and anomaly segmentation methods. Specifically, we achieved improvements over the state-of-the-arts by approximately 6\% AUPR-Error on Cityscapes pixel-level error prediction, 7\% Pearson correlation on pancreatic tumor DSC prediction and 20\% AUPR on StreetHazards anomaly segmentation.

We summarize our contribution as follows:
\begin{itemize}
    \item To the best of our knowledge, we are the first to systematically study failure detection and anomaly detection for semantic segmentation 
    \item We propose a unified framework, SynthCP, which enjoys the benefits of a semantic-to-image conditional GAN, to address both of the two tasks. 
    \item SynthCP achieves state-of-the-art failure detection and anomaly segmentation results on three challenging datasets. 
\end{itemize}

\section{Related Work}
In this section, we first review the topics closely related to failure detection and anomaly segmentation, such as 
uncertainty/confidence estimation, quality assessment and out-of-distribution (OOD) detection. Then, we review generative adversarial networks (GANs), which serves a key module in our framework.

\noindent\textbf{Uncertainty estimation} or confidence estimation has been a hot topic in the field of machine learning for years, and can be directly applied to the task of \textbf{failure detection}.
Standard baselines was established in~\cite{hendrycks2016baseline} for detecting failures in classification where maximum softmax probability (MSP) provides reasonable results. However, the main drawback of using MSP for confidence estimation is that deep networks tend to produce high confidence predictions~\cite{tcp}. Geifman \textit{et al.}~\cite{conf4} controled the user specified risk-level by setting up thresholds on a pre-defined confidence function (e.g. MSP). Jiang \textit{et al.}~\cite{trustscore} measured the agreement between the classifier and a modified nearest-neighbor classifier on the test examples as a confidence score. A recent approach~\cite{tcp} proposed to direct regress ``true class probability'' which improved over MSP for failure detection. Additionally, Bayesian approaches have drawn attention in this field of study. Dropout based approaches ~\cite{gal2016dropout, Unc17} used Monte Carlo Dropout (MCDropout) for Bayesian approximation. Computing statistics such as entropy or variance is capable of indicating uncertainty. However, all these approaches mainly focus on small image classification tasks. When applied to semantic segmentation, they lack the information of semantic structures and contexts. 
 
\noindent
\textbf{Segmentation quality assessment} aims at estimating the overall quality of segmentation, without using ground-truth label, which is suitable to make alarms when model fails. Some approaches~\cite{kwon2020uncertainty,SNC18} utilize Bayesian CNNs to predict the segmentation quality of medical images. \cite{robinson2018real, EWOG12} regressed the segmentation quality from deep features computed from a pair of an image and its segmentation result. \cite{iounet1,iounet2} plugged an extra IoU regression head into object detection or instance segmentation. \cite{chabrier2006unsupervised, gao2017novel} used unsupervised learning methods to estimate the segmentation quality using geometrical features. Recently, Liu \textit{et al.}~\cite{liu2019alarm} proposed to use VAE~\cite{vae} to capture a shape prior for segmentation quality assessment on 3D medical image. However, it is hardly applicable to natural images considering the complexity and large shape variance in 2D scenes and objects. Segmentation quality assessment will be referred to as image-level failure detection in the rest of the paper.

\noindent
\textbf{OOD detection} aims at detecting out-of-distribution examples in testing data. Since the baseline MSP method~\cite{hendrycks2016baseline} was brought up, many approaches have improved OOD detection from various aspects~\cite{ood1,ood2,ood3,ood4,ood5}. While these approaches mainly focus on image level OOD detection, \emph{i.e.}, to determine whether an image is an OOD example, \emph{e.g.}, \cite{wilddash1, wilddash2} targeted at detecting hazardous scenes in the Wilddash dataset~\cite{zendel2018wilddash}. On the contrary, we focus on \textbf{anomaly segmentation}, \emph{i.e.}, a pixel-level OOD detection task that aims at segmenting anomalous regions from an image. Pixel-wise reconstruction loss~\cite{baur2018medical, haselmann2018product} with auto-encoders(AE) are the main stream approaches for anomaly segmentation. However, they can hardly model the complex street scenes in natural images and AEs can not guarantee to generate an in-distribution image from OOD regions. Recently, it was found that MSP surprisingly outperform AE and Bayesian network based approaches on a newly built larger scale street scene dataset StreetHazards~\cite{caos} - with 250 types of anomalous objects and more than 6k high resolution images.  Lis \textit{et al.}~\cite{lis2019detecting} proposed to re-synthesize an image from the predicted semantic map to detect OOD objects in street scenes, which is the \textbf{pioneer} work for synthesis-based anomaly detection for semantic segmentation. SynthCP also follows this spirit, but we use a simple yet effective feature distance measure rather than a discrepancy network to find anomalies. In addition, we extend this idea to do a systematic study of failure detection.

\noindent
\textbf{Generative adversarial networks}~\cite{gan} generate realistic images by playing a ``min-max'' game between a generator and a discriminator. GANs effectively minimize a Jensen-Shannon divergence, thus generating in-distribution images. SynthCP utilizes conditional GANs~\cite{cGAN} (cGANs) for image translation~\cite{pix2pix}, \emph{a.k.a} pixel-to-pixel translation. Approaches designed for semantic image synthesis~\cite{pix2pixhd, spade, liu2019learning} improves pixel-to-pixel translation in synthesizing real images from semantic masks, which is the reverse procedure of semantic segmentation. Since semantic image synthesis is commonly evaluated by the performance of a segmentation model, reversely, we are motivated to use a semantic-to-image generator for failure detection for semantic segmentation.

\begin{figure}[!t]
\begin{center}
    \includegraphics[width=\textwidth]{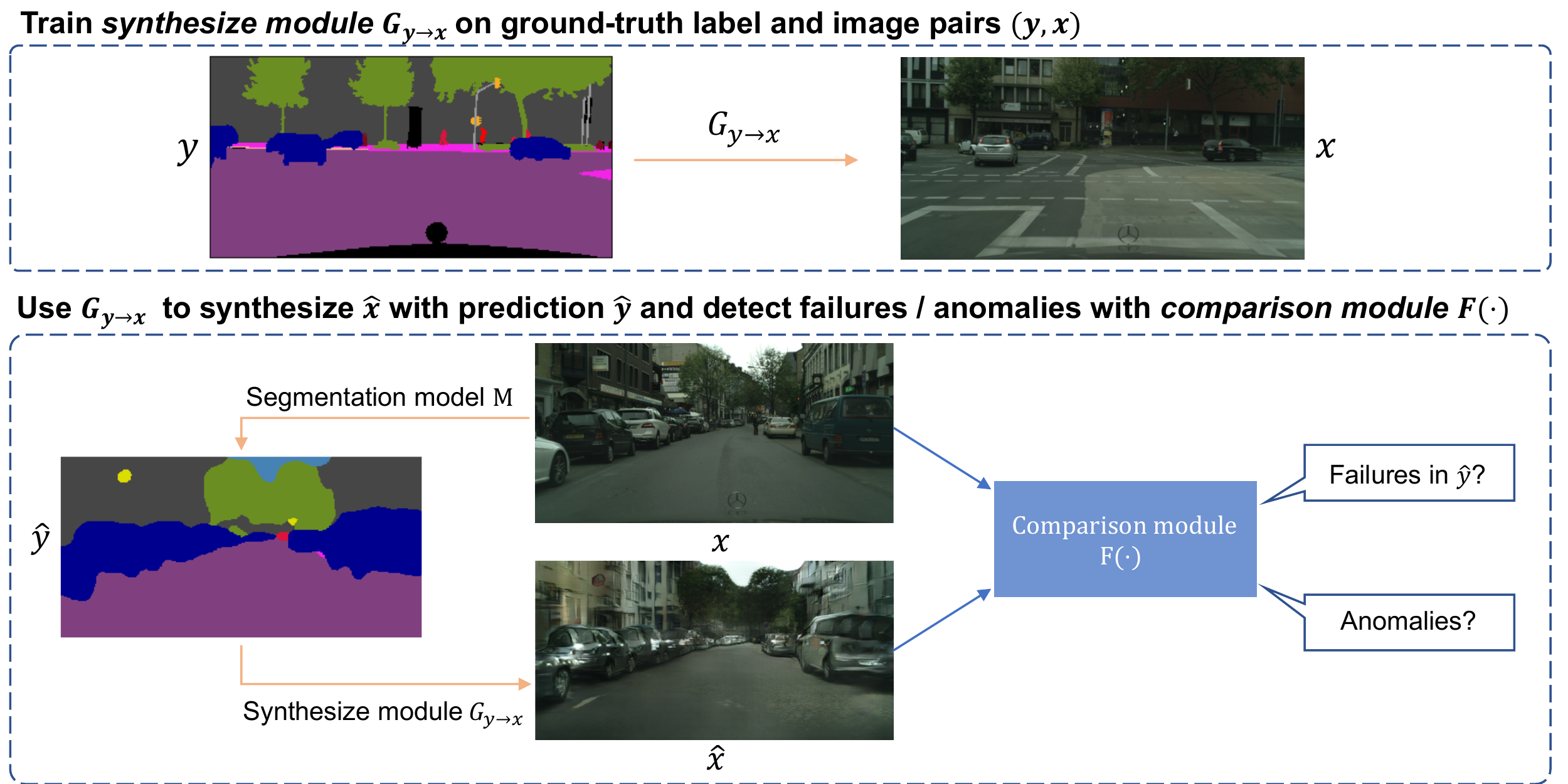}
\end{center}
\caption{
   We first train the synthesis module $G_{y\rightarrow x}$ on label-image pairs and then use this module to synthesize the image conditioning on the predicted segmentation mask $\hat{y}$. By comparing $x$ and $\hat{x}$ with a comparison module $F(\cdot)$, we can detect failures as well as segment anomalous objects. $F(\cdot)$ is instantiated in Sec~\ref{sec4.1} and Sec~\ref{sec5.1}.
}
\label{Fig2}
\end{figure}

\section{Methodology}
In this section, we introduce our framework, SynthCP, for failure detection and anomaly detection for semantic segmentation. SynthCP consists of two modules, an image synthesis module and a comparison  module. We first introduce the general framework (shown in Fig.~\ref{Fig2}), then describe the details of the modules for failure detection and anomaly detection in Sec~\ref{sec4} and Sec~\ref{sec5}, respectively. Unless otherwise specified, the notations in this paper follow this criterion: We use a lowercase letter, \emph{e.g.}, $x$, to represent a tensor variable, such as a 1D array or a 2D map, and denote its $i$-th element as $x^{(i)}$; We use a capital letter, \emph{e.g.}, $F$, to represent a function.



\subsection{General Framework}
Let $x$ be an image with size of $w \times h$ and $\mathbb{L}=\{1,2,\ldots,L\}$ be a set of integers representing the semantic labels. By feeding image $x$ to a segmentation model $M$, we obtain its segmentation result, \emph{i.e.}, a pixel-wise semantic label map $\hat{y} = M(x) \in \mathbb{L}^{w\times h}$. Our goal is to identify and locate the failures in $\hat{y}$ or detect anomalies in $x$ based on $\hat{y}$. 

\subsubsection{Image Synthesis Module}
We model this image synthesize module by a pixel-to-pixel translation conditional GAN (cGAN)~\cite{cGAN}, which is known for its excellent ability for semantic-to-image mapping. It consists of a generator $G$ and a discriminator $D$. 

\noindent
\underline{Training.} We train this translation conditional GAN on label-image pairs: $(y,x)$, where $y$ is a grouth-truth pixel-wise  semantic label map and $x$ is its corresponding image. The objective of the generator $G$ is to translate semantic label maps to realistic-looking images, while the discriminator $D$ aims to distinguish real images from the synthesized ones. This cGAN minimizes the conditional distribution of real images via the following min-max game:

\begin{equation}
    \min_G \max_D \mathcal{L}_{GAN}(G, D),
\end{equation}
where the objective function $\mathcal{L}_{GAN}(G, D)$ is defined as:
\begin{equation}
    \mathbb{E}_{({y},x)}[\log D({y}, x)] + \mathbb{E}_{{y}}[\log(1 - D({y}, G({y}))].
\end{equation}

\noindent
\underline{Testing.} After training, we fix the generator $G$. Given an image $x$ and a segmentation model $M$, we feed the predicted segmentation mask $\hat{y}=M(x)$ into $G$, and obtain a synthesized (\emph{i.e.}, reconstructed) image $\hat{x}$: 
\begin{equation}
    \hat{x} = G(\hat{y}).
\end{equation}
$\hat{x}$ and $x$ are then served as the input for the comparison module. 

\subsubsection{Comparison Module}
We detect failures and anomalies in $\hat{y}$ by comparing $\hat{x}$ with $x$. Our assumption is that, if $\hat{x}$ is more similar to $x$, then $\hat{y}$ is more similar to $y$. However, since the optimization of $G$ does not guarantee that the synthesized image $\hat{x}$ has the same style as the original image $x$, simple similarity measurements such as $\ell_1$ distance between $x$ and $\hat{x}$ is not accurate. In order to address this issue, we model the comparison module by a task-specific function $F$ which estimates a trustworthy task-specific confidence measure $\hat{c}$ between $x$ and $\hat{x}$:
\begin{equation}
    \hat{c} = F(x, \hat{x}) = F(x, G(\hat{y})).
\end{equation}

For the task of failure detection, the confidence measure $\hat{c} = (\hat{c}_{iu}, \hat{c}_m)$ includes an image-level per-class intersection over union (IoU) array $\hat{c}_{iu}\in [0,1]^{|\mathbb{L}|}$ and a pixel-level error map $\hat{c}_m\in[0,1]^{w\times h}$; For the task of anomaly segmentation, the confidence measure $\hat{c}$ is a pixel-level confidence map $\hat{c}_n\in[0,1]^{w\times h}$ for anomalous objects.

\subsection{Failure Detection}
\label{sec4}
\subsubsection{Problem Definition} Our failure detection contains two tasks: 1) a per-class IoU prediction $\hat{c}_{iu} \in [0,1]^{|\mathbb{L}|}$, which is useful to indicate whether there are failures in the segmentation result $\hat{y}$, and 2) to locate the failures in $\hat{y}$, which needs to compute a  pixel-level error map $\hat{c}_m \in [0,1]^{w\times h}$. 



\subsubsection{Instantiation of Comparison Module}

\begin{figure}[!t]

\begin{center}
    \includegraphics[width=1.0\textwidth]{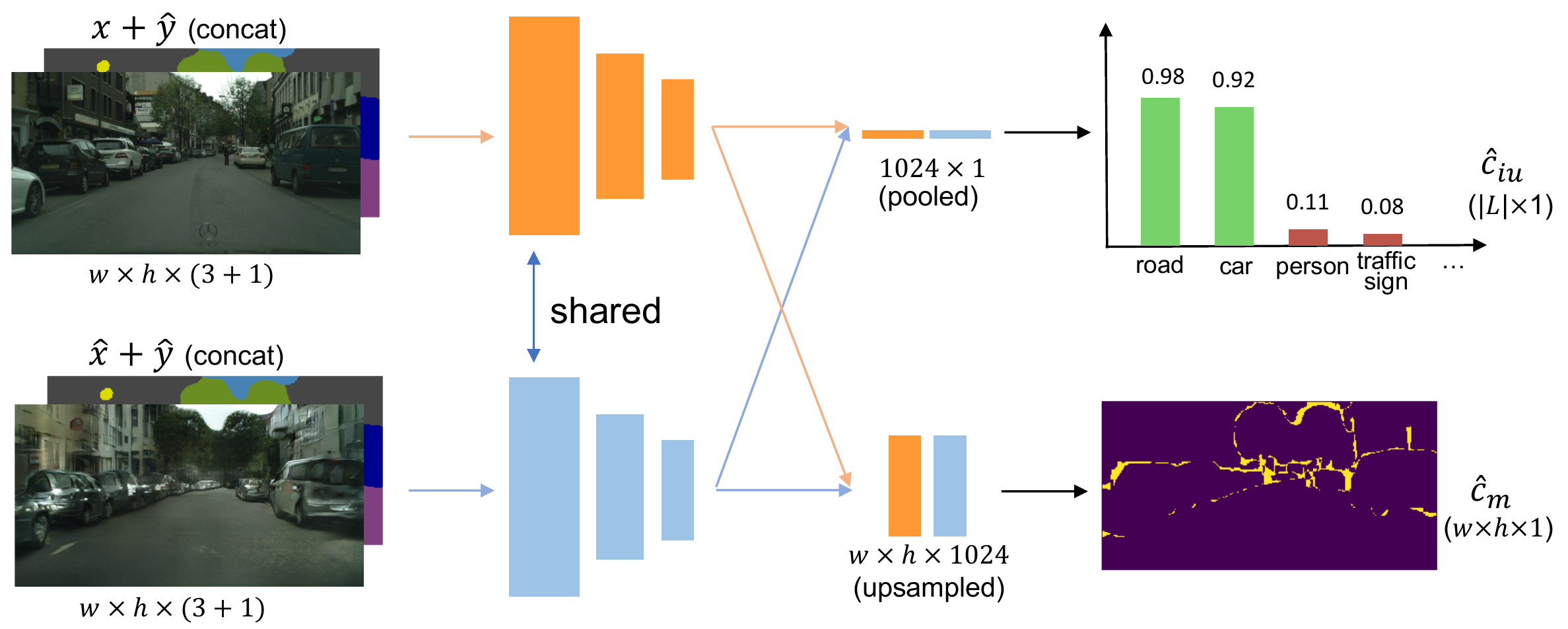}
\end{center}
\caption{
   We instantiate $F(\cdot)$ as a light-weighted siamese network $F(x, \hat{x}, \hat{y};\theta)$ for joint image-level per-class IoU prediction and pixel-level error map prediction.
}
\label{Fig3}
\end{figure}

\label{sec4.1} We instantiate the comparison module $F(\cdot)$ as a light-weighted deep network. In practice, we use ResNet-18~\cite{he2016resnet} as the base network and follow a siamese-style design for learning the relationship between $x$ and $\hat{x}$. As illustrated in Fig.~\ref{Fig3}, $x$ and $\hat{x}$ are first concatenated with $\hat{y}$ and then separately encoded by a shared-weight siamese encoder. Then two heads are built upon the siamese encoder and output the image-level per-class IoU array $\hat{c}_{iu} \in [0,1]^{|\mathbb{L}|}$ and pixel-level error map $\hat{c}_m \in \left[0,1\right]^{w\times h} $, respectively. We rewrite the function $F$ for failure detection as below:

\begin{equation}
    \hat{c}_{iu}, \hat{c}_m = F(x, \hat{x}, \hat{y}; \theta)
\end{equation}
where $\theta$ represents the network parameters.

In the training stage, the supervision of network training is obtained by computing the ground-truth confidence measure $c$ from $y$ and $\hat{y}$. For the ground-truth image-level per-class IoU array ${c}_{iu}$, we compute it by 
\begin{equation}
{c}_{iu}^{(l)}=\frac{|\{i|\hat{y}^{(i)}=l\}\cap \{i|{y}^{(i)}=l\}|}{|\{i|\hat{y}^{(i)}=l\}\cup \{i|{y}^{(i)}=l\}|},
\end{equation}
where $l$ is the $l$-th semantic class in label set $\mathbb{L}$. The $\ell_1$ loss function $\mathcal{L}_{\ell_1}({c}_{iu}^{(l)}, \hat{c}_{iu}^{(l)})$ is applied to learning this image-level per-class IoU prediction head. For the ground-truth pixel-level error map, we compute it by
\begin{equation}
c_m^{(i)}=\left\{
\begin{aligned}
&1 &\text{if} \; y^{(i)} \neq \hat{y}^{(i)}\\
&0 &\text{if} \; y^{(i)} = \hat{y}^{(i)} \\
\end{aligned}
\right..
\end{equation}
The binary cross-entropy loss $\mathcal{L}_{ce}(c_m^{(i)},\hat{c}_m^{(i)})$ is applied to learning this pixel-level error map prediction head. The overall loss function of failure detection $\mathcal{L}$ is the sum of the above two:

\begin{equation}
    \mathcal{L} = \frac{1}{|\mathbb{L}|} \sum_l^{|\mathbb{L}|}\mathcal{L}_{\ell_1}({c}_{iu}^{(l)}, \hat{c}_{iu}^{(l)}) + \frac{1}{wh}\sum_i^{wh} \mathcal{L}_{ce}(c_m^{(i)},\hat{c}_m^{(i)}).
\end{equation}

\subsection{Anomaly Segmentation}
\label{sec5}

\subsubsection{Problem Definition} The goal of anomaly segmentation is segmenting anomalous objects in a test image which are unseen in the training images. Formally, given a test image $x$, an anomaly segmentation method should output a confidence score map $\hat{c}_n \in [0,1]^{w\times h}$ for the regions of the anomalous objects in the image, \emph{i.e.},  $\hat{c}_n^{(i)} = 1$ and $\hat{c}_n^{(i)} = 0$ indicate the $i^{th}$ pixel belongs to an anomalous object and an in-distribution object (the object is seen in the training images), respectively.

\subsubsection{Instantiation of Comparison Module}
\label{sec5.1}
As the same as failure detection, we first train a cGAN generator $G$ on the training images, which maps the in-distribution object labels to realistic images. Given a semantic segmentation model $M$, we feed its prediction $\hat{y}=M(x)$ into $G$ and obtain $\hat{x} = G(\hat{y})$. Since $\hat{y}$ only contains in-distribution object labels, $\hat{x}$ also only contain in-distribution objects. Thus, we can compare $x$ with $\hat{x}$ to find the anomalies. The pixel-wise semantic difference of $x$ and $\hat{x}$ is a strong indicator of anomalous objects. Here, we simply instantiate the comparison function $F(\cdot)$ as the cosine distance defined on the intermediate features extracted by the segmentation model $M$:

\begin{equation}
    \hat{c}_n^{(i)} = F(x, \hat{x};M) = 1 - \langle \frac{\mathbf{f}_M^i(x)}{\lVert \mathbf{f}_M^i(x)\rVert_2}, \frac{\mathbf{f}_M^i(\hat{x})}{\lVert \mathbf{f}_M^i(\hat{x}) \rVert_2} \rangle
\end{equation}
where $\mathbf{f}_M^i$ is the feature vector at the $i^{th}$ pixel position outputted by the last layer of segmentation model $M$ and $\langle\cdot,\cdot\rangle$ is the inner product of the two vectors.

\noindent
\underline{Post-processing with MSP.} Due to the artifacts and generalized errors of GANs, our approach may mis-classify an in-distribution object into an anomalous object (false positives). We use a simple post-processing to address this issue. We refine the result by maximum softmax probability (MSP)~\cite{hendrycks2016baseline}, which is known as an effective uncertainty estimation strategy: $\hat{c}_n^{(i)} \leftarrow \hat{c}_n^{(i)} \cdot \mathbbm{1}\{p^{(i)} \leq t \} + (1-p^{(i)}) \cdot \mathbbm{1}\{p^{(i)} > t\}$, where $p^{(i)}$ is the maximum soft-max probability at the $i$-th pixel outputted by the segmentation model $M$,  $t \in [0,1]$ is a threshold and $\mathbbm{1}\{\cdot\}$ is the indicator function.



\begin{figure}[!b]
\begin{center}
    \includegraphics[width=\textwidth]{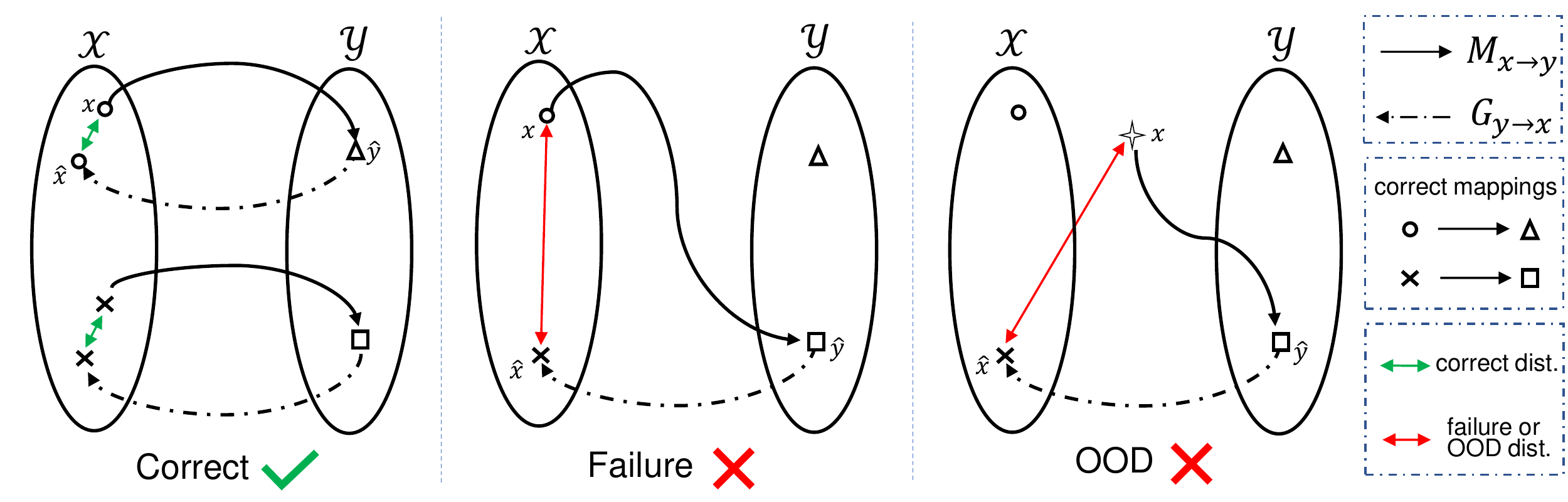}
\end{center}
\caption{
   An analysis of SynthCP. Left: $M_{x\rightarrow y}$ correctly maps $x$ to $\hat{y}$, resulting in small distance between $x$ and the synthesized $\hat{x}$. However, when there are failures in $\hat{y}$ (middle) or there are OOD examples in $x$ (right), the distance between $x$ and $\hat{x}$ is larger, given a reliable reverse mapping $G_{y\rightarrow x}$.
}
\label{Fig:analysis}
\end{figure}

\subsection{Conceptual Explanation}

We give conceptual explanations of SynthCP in Fig.~\ref{Fig:analysis}, where $\mathcal{X}$ and $\mathcal{Y}$ correspond to image space and label space. $M_{x\rightarrow y}$ is the segmentation model and $G_{y\rightarrow x}$ is a semantic-to-image generator. The left image shows when $M_{x\rightarrow y}$ correctly maps an image to its corresponding segmentation mask, the synthesized image generated from $G_{y\rightarrow x}$ is close to the original image. However, when $M_{x\rightarrow y}$ makes a failure (middle) or encounters an OOD case (right), the synthesized image should be far away from the original image. As a result, the synthesized image serves as a strong indicator for either failure detection or OOD detection.

\section{Experiments}
\subsection{Failure Detection}
\subsubsection{Evaluation Metrics} Following~\cite{liu2019alarm}, we evaluate the performance of image-level failure detection, \emph{i.e.}, per-class IoU prediction, by four metrics:  \textbf{MAE}, \textbf{STD}, \textbf{P.C} and \textbf{S.C}. MAE (mean absolute error) and its STD measure the average error between predicted IoUs and ground-truth IoUs. P.C (Pearson correlation) and S.C.(Spearman correlation) measures their correlation coefficients. For pixel-level failure detection, \emph{i.e.}, pixel-level error map prediction, we use the metrics in literature~\cite{hendrycks2016baseline, tcp}: \textbf{AUPR-Error}, \textbf{AUPR-Success}, \textbf{FPR at 95\% TPR} and \textbf{AUROC}. Following~\cite{tcp}, AUPR-Error is our main metric, which computes the area under the Precision-Recall curve using errors as the positive class.  

\begin{table}[!b]
    \centering
 \caption{Experiments on the Cityscapes dataset. We detect failures in the segmentation results of FCN-8 and Deeplab-v2. ``SynthCP-separate" and ``SynthCP-joint" mean training the image-level and pixel-level failure detection heads in our network separately and jointly, respectively.}
    \label{tab:cityscapes}
    \begin{tabular}{l c c c c c c c c c }
\hline
\multicolumn{1}{l|}{} 
& \multicolumn{4}{c}{FCN-8 }& \multicolumn{4}{|c}{Deeplab-v2} \\
\multicolumn{1}{l|}{image-level}  & MAE$\downarrow$ & STD$\downarrow$ & P.C.$\uparrow$ & \multicolumn{1}{c|}{S.C.$\uparrow$}  & MAE$\downarrow$ & STD$\downarrow$ & P.C.$\uparrow$& \multicolumn{1}{c}{S.C.$\uparrow$} \\
\hline
\multicolumn{1}{l|}{MCDropout~\cite{gal2016dropout}}& 17.28&	13.33&	3.62&	\multicolumn{1}{c|}{5.97}
& 19.31&	12.86&	4.55&	\multicolumn{1}{c}{1.37}\\
\multicolumn{1}{l|}{VAE alarm~\cite{liu2019alarm}}& 16.28&	11.88&	21.82&	\multicolumn{1}{c|}{18.26}
& 16.78&	12.21&	17.92&	\multicolumn{1}{c}{19.63}\\
\multicolumn{1}{l|}{Direct Prediction} & 13.25&	11.96&	58.34&	\multicolumn{1}{c|}{59.74}
& 14.45&	12.20&	60.94&	\multicolumn{1}{c}{62.01}\\
\hline
\multicolumn{1}{l|}{SynthCP-separate} & \textbf{11.58}&	11.50&	\textbf{64.63}&	\multicolumn{1}{c|}{\textbf{65.63}}
& \textbf{13.60}&	12.32 &	62.51 &	\multicolumn{1}{c}{63.41}\\
\multicolumn{1}{l|}{SynthCP-joint} & 12.69&	\textbf{11.29}&	62.52&	\multicolumn{1}{c|}{61.23}
& 13.68&	\textbf{11.60}&	\textbf{64.05} &	\multicolumn{1}{c}{\textbf{65.42}}\\
\hline\hline
\multicolumn{1}{l|}{} 
& \multicolumn{4}{c}{FCN-8 }& \multicolumn{4}{|c}{ Deeplab-v2 } \\
\multicolumn{1}{l|}{pixel-level}  & AP-Err$\uparrow$ & AP-Suc$\uparrow$ & AUC$\uparrow$ & \multicolumn{1}{c|}{FPR95$\downarrow$}  & AP-Err$\uparrow$ & AP-Suc$\uparrow$ & AUC$\uparrow$ & \multicolumn{1}{c}{FPR95$\downarrow$} \\
\hline
\multicolumn{1}{l|}{MSP~\cite{hendrycks2016baseline}} & 50.31&	99.02&	91.54&	\multicolumn{1}{c|}{25.34} 
& 48.46&	99.24&	92.26&	\multicolumn{1}{c}{24.41} \\
\multicolumn{1}{l|}{MCDropout~\cite{gal2016dropout}} & 49.23&	99.02&	91.47&	\multicolumn{1}{c|}{25.16}
 & 47.85&	99.23&	92.19&	\multicolumn{1}{c}{24.68}\\
\multicolumn{1}{l|}{TCP~\cite{tcp}} & 48.54&	98.82&	90.29&	\multicolumn{1}{c|}{32.21}
 & 45.57&	98.84&	89.14&	\multicolumn{1}{c}{36.98}\\
\multicolumn{1}{l|}{Direct Prediction} & 52.17&	99.15&	92.55&	\multicolumn{1}{c|}{\textbf{22.34}} 
& 48.76&	99.34&	92.94&	\multicolumn{1}{c}{\textbf{21.56}}\\
\hline
\multicolumn{1}{l|}{SynthCP-separate} & 54.14&	99.15&	92.70&	\multicolumn{1}{c|}{22.47}
& 48.79 &	99.31 &	92.74 &	\multicolumn{1}{c}{22.15}\\
\multicolumn{1}{l|}{SynthCP-joint} & \textbf{55.53}&	\textbf{99.18}&	\textbf{92.92}&	\multicolumn{1}{c|}{22.47}
& \textbf{49.99} &	\textbf{99.34}&	\textbf{92.98}&	\multicolumn{1}{c}{21.69}\\
\hline
\end{tabular}
   
\end{table}

\subsubsection{The Cityscapes Dataset}
We validate SynthCP on the Cityscapes dataset~\cite{cordts2016cityscapes}, which contains 2975 high-resolution training images and 500 validation images. As far as we know, it is the largest one for failure detection for semantic segmentation.

\noindent
\underline{Baselines.} We compare SynthCP to MCDropout~\cite{gal2016dropout}, VAE alarm~\cite{liu2019alarm}, MSP~\cite{hendrycks2016baseline}, TCP~\cite{tcp} and ``Direct Prediction". MCDropout, MSP and TCP output pixel-level confidence maps, serving as standard baselines for pixel-level failure prediction. VAE alarm~\cite{liu2019alarm} is the state-of-the-art in image-level failure prediction method. Following~\cite{liu2019alarm}, we also use MCDropout to predict image-level failures. Direct Prediction is a method that directly uses a network to predict both image-level and pixel-level failures, by taking an image and its segmentation result as input. Note that, Direct Prediction shares the same experimental settings (backbone and training strategies) with SynthCP, which can be seen as an ablation study on the effectiveness of the synthesized image $\hat{x}$.


\noindent
\underline{Implementation details.} We use the state-of-the-art semantic-to-image cGAN - SPADE~\cite{spade} in SynthCP. 
We re-trained SPADE from scratch following the same hyper-parameters as in~\cite{spade} with only semantic segmentation maps as the input (without the instance maps). The backbone of our comparison module is ResNet-18~\cite{he2016resnet} pretrained from Image-Net. We use ImageNet pre-trained model and train the network for 20k iterations using Adam optimizer~\cite{kingma2014adam} with initial learning 0.01 and $\beta = (0.9,0.999)$, which takes about 6 hours on one single Nvidia Titan Xp GPU.
Since we use a network to predict failures, we need to generate training data for this network. A straightforward strategy is to divide the original training set into a training subset and a validation subset, then train the segmentation model on the training subset and test it on the validation subset. The testing results on the validation subset can be used to train the failure predictor. We extend this strategy by doing 4-fold cross validation on the training set. Since the cross-validated results cover all samples in the original training set, we are able to generate sufficient training data to train our failure prediction network.

\begin{figure}[!t]
\begin{center}
    \includegraphics[width=\textwidth]{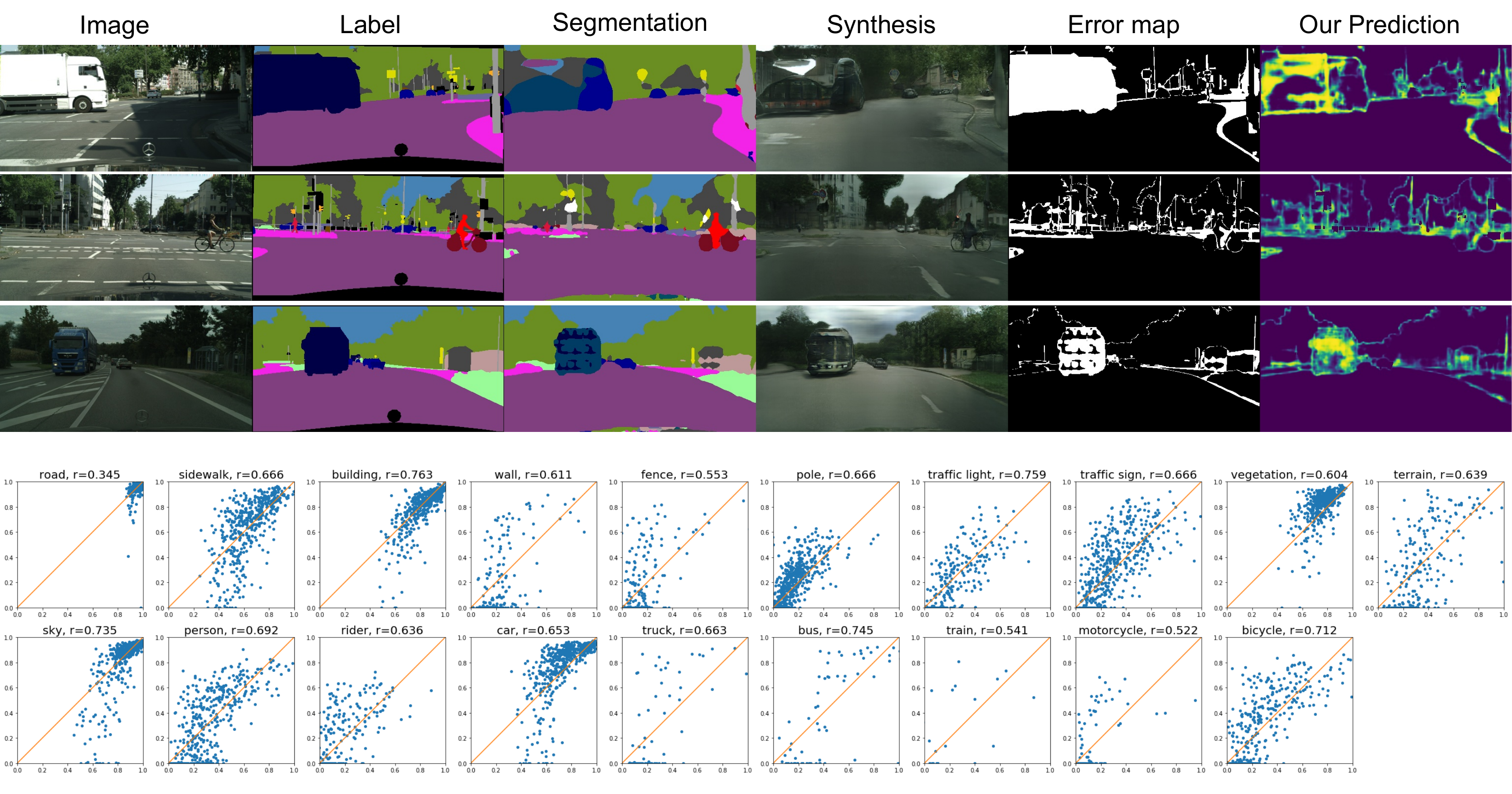}
\end{center}
\caption{
   Visualization on the Cityscapes dataset for pixel-level error map prediction (top) and image-level per-class IoU prediction (bottom). For each example from left to right (top), we show the original image, ground-truth label map, segmentation prediction, synthesized image conditioned on the segmentation prediction, (ground-truth) errors in the segmentation prediction and our pixel-level error prediction. The plots (bottom) show significant correlations between the ground-truth IoU and our predicted IoU on most of the classes.
}
\label{Fig5}
\end{figure}

\noindent
\underline{Results.} Experimental results are shown in Table~\ref{tab:cityscapes} and visualizations are shown in Fig.~\ref{Fig5}.
We use the well-known FCN8~\cite{long2015fcn} and Deeplab-v2~\cite{deeplab} as the segmentation models. For image-level failure detection, our approach consistently outperforms other methods on all metrics. Results are averaged over 19 classes for all four metrics (detailed results in supplementary). We find that VAE alarm does not perform well 2D images of street scenes, since small objects are easily missed in the VAE reconstruction. Without the synthesized images from the segmentation results, Direct Prediction performs worse than ours despite achieving better performance than the others.

For pixel-level failure detection, our approach achieves the state-of-the-art performance as well, especially for AP-Error metric where our approach outperforms other methods by a considerable margin. The comparison to Direct Prediction demonstrates that the improvements come from the image synthesis module in our framework. We hypothesis that TCP performs not as well because it is mainly designed for classification and it might be hard to fit the true class probability for dense predictions on large images in our settings.
We find that our method produces slightly more false positives than ``Direct Prediction" baseline (FPR95 is lower). We think the reason might be some correctly segmented regions are not synthesized well by the generative model. 

We conducted another experiment to validate the \textbf{generalizability} on unseen segmentation models. We directly test our failure detection model, which is trained on Deeplab-v2 masks, on the segmentation masks produced by FCN8. We achieve an AUPR-Error of 53.12 for pixel-level error detection and MAE of 12.91 for image-level failure prediction. Full results are available in the supplementary material. The results are comparable to those obtained by our model trained on FCN8 segmentation model, as shown in table~\ref{tab:cityscapes}.

\begin{table}[!b]
    \centering
 \caption{Failure detection results on the pancreatic tumor segmentation dataset in MSD~\cite{simpson2019msd}}
    \label{tab:msd}
    \begin{tabular}{l p{0.1\textwidth} p{0.1\textwidth} p{0.1\textwidth} p{0.1\textwidth} p{0.1\textwidth} }
\hline
\multicolumn{1}{l|}{} 
& \multicolumn{4}{c}{tumor DSC prediction} \\
\multicolumn{1}{l|}{Method}  &  MAE $\downarrow$ & STD $\downarrow$ & P.C. $\uparrow$ & S.C. $\uparrow$\\
\hline\hline
\multicolumn{1}{l|}{Direct Prediction} & 23.20 &29.81 &45.50& 45.36\\
\multicolumn{1}{l|}{Jungo \textit{et al.} \cite{SNC18}} & 26.57 & 29.78 &-23.87 &-20.23\\
\multicolumn{1}{l|}{Kwon \textit{et al.} \cite{kwon2020uncertainty}} & 26.14 &29.24 &14.61 &14.70\\
\multicolumn{1}{l|}{VAE alarm~\cite{liu2019alarm}} & 20.21 &23.60 &60.24 &63.30\\
\multicolumn{1}{l|}{VAE (our imple.)} & 18.60& 13.73&	63.42&	58.47\\
\hline
\multicolumn{1}{l|}{SynthCP} & 18.13&	13.77&	61.11&	62.66\\
\multicolumn{1}{l|}{SynthCP + VAE} & \textbf{15.19} & \textbf{13.37} & \textbf{67.97}	& \textbf{71.35}\\
\hline

\end{tabular}
\label{tab:msd}
\end{table}

\subsubsection{The Pancreatic Tumor Segmentation Dataset}
\label{sec:tumor}
We also validate SynthCP on medical images. Following VAE alarm~\cite{liu2019alarm}, we applied SynthCP to the challenging pancreatic tumor segmentation task of Medical Segmentation Decathlon~\cite{simpson2019msd}, where we randomly split the 281 cases into 200 training and 81 testing. The VAE alarm system~\cite{liu2019alarm} is the main competitor on this dataset. Since their approach explored shape prior for accurate quality assessment and tumor shapes have large variance, we expect SynthCP can outperform shape-based models or be complementary to the VAE-based alarm model. We only compare image-level failure detection in this dataset, because the VAE alarm system~\cite{liu2019alarm} is targeted to this task and sets up standard baselines.

\begin{figure}[!t]
\begin{center}
    \includegraphics[width=\textwidth]{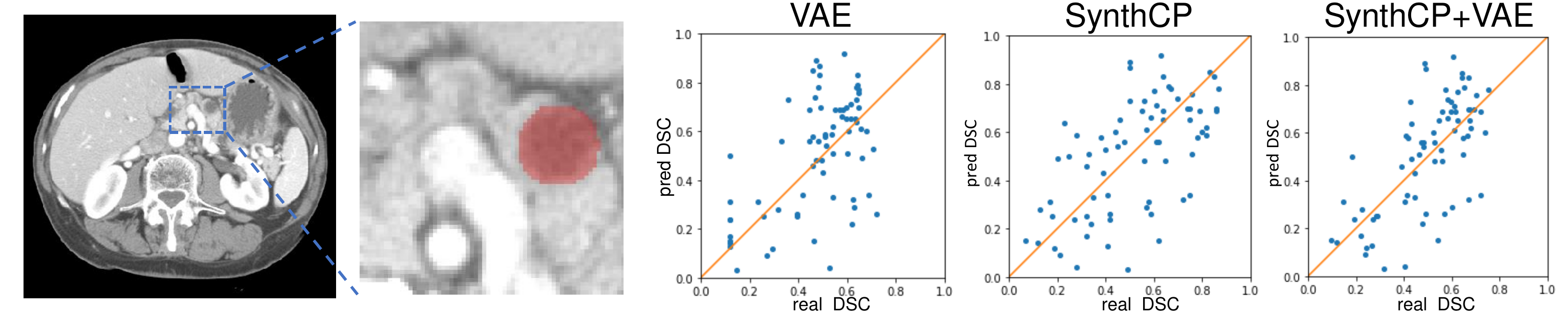}
\end{center}
\caption{
   Left two: an example of pancreatic tumor segmentation (in red). Right three: plots for tumor segmentation DSC score prediction by VAE alarm~\cite{liu2019alarm}, SynthCP and the ensemble of SynthCP and VAE alarm.
}
\label{Fig:vis_msd}
\end{figure}

We use the state-of-the-art network 3D AH-Net~\cite{ahnet} as the segmentation model. Instead of IoU, the segmentation performance is measured by Dice coefficient (DSC), a standard evaluation metric used for medical image segmentation. Moving into 3D is challenging for SynthCP, since training 3D GANs is extremely hard, considering the limited GPU memory and high computational costs. In practice, we modify SPADE into 3D. Results and visualizations are shown in Table~\ref{tab:msd} and Fig.~\ref{Fig:vis_msd} respectively. In terms of baselines, we re-implement the VAE alarm system for a fair comparison in our settings, while the results of other methods are quoted  from~\cite{liu2019alarm}. SynthCP achieved comparable performances as VAE alarm system. When combined with VAE alarm (a simple ensemble of the predicted DSC), all of the four metric improves significantly (P.C. and S.C correlation coefficient both improves by approximately 7\% and 10\% respectively), illustrating SynthCP which captures label-to-image information is complementary to the shape-based VAE approach.

\subsection{Anomaly Segmentation}
\subsubsection{Evaluation metrics.} We use the standard metrics for OOD detection and anomaly segmentation:  area under the ROC curve (AUROC), false positive rate at 95\% recall (FPR95), and area under the precision recall curve (AUPR). 

\subsubsection{The StreetHazards Dataset}

\begin{table}[!b]
\centering
\caption{
    Anomaly segmentation results on StreetHazards dataset~\cite{caos}
}
\begin{tabular}{l|c c c c}    
\hline
Method & FPR95$\downarrow$ & AUROC$\uparrow$ & AUPR$\uparrow$\\
\hline\hline
AE~\cite{baur2018medical}  & 91.7 & 66.1 & 2.2 \\
Dropout~\cite{gal2016dropout}  & 79.4  & 69.9 & 7.5 \\
MSP~\cite{hendrycks2016baseline}& 33.7  & 87.7 & 6.6 \\
MSP + CRF~\cite{caos}& 29.9  & 88.1 & 6.5\\
\hline
SynthCP  &  \textbf{28.4} &\textbf{88.5} & \textbf{9.3}\\
\hline
\end{tabular}
\label{tab:caos}
\end{table}

We validate SynthCP on the StreetHazards dataset of CAOS Benchmark~\cite{caos}. This dataset contains 5125 training images, 1000 validation images and 1500 test images. 250 types of anomaly objects appears only in the testing images.

\noindent
\underline{Baselines.} Baseline approaches include MSP~\cite{hendrycks2016baseline}, MSP+CRF~\cite{caos}, Dropout~\cite{gal2016dropout} and an auto-encoder (AE) based approach~\cite{baur2018medical}. Except for AE, all the other three approaches require a segmentation model to provide either softmax probability or uncertainty estimation. AE is the only approach that requires extra training of an auto-encoder for the images and computes pixel-wise $\ell_1$ loss for anomaly segmentation.

\noindent
\underline{Implementation details.} Following~\cite{caos},  we use two network backbones as the segmentation models: ResNet-101~\cite{he2016resnet} and PSPNet~\cite{zhao2017psp}. The cGAN is also SPADE~\cite{spade} trained with the same training strategy as in Sec~\ref{sec4}. The post-processing threshold $t = 0.999$ is chosen for better AUPR and is discussed in detail in the following paragraph.

\begin{figure}[!t]
\begin{center}
    \includegraphics[width=\textwidth]{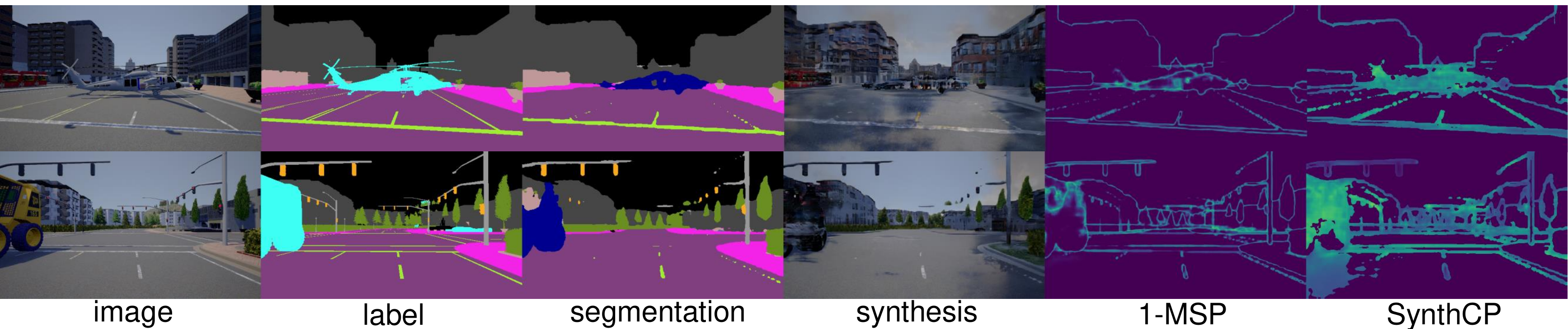}
\end{center}
\caption{
   Visualizations on the StreetHazards dataset. For each example, from left to right, we show the original image, ground-truth label map, segmentation prediction, synthesized image conditioned on segmentation prediction, MSP anomaly segmentation prediction and our anomaly segmentation prediction.
}
\label{vis:caos}
\end{figure}

\noindent
\underline{Results} Experimental results are shown in Table~\ref{tab:caos}. SynthCP improves the previous state-of-the-art approach MSP+CRF from 6.5\% to 9.3\% in terms of AUPR. Fig.~\ref{vis:caos} shows some anomaly segmentation examples. 

To study how much MSP post-processing contributes to SynthCP, we conduct experiments on different thresholds of $t$ for post-processing. As shown in Table~\ref{tab:th_post_process}, without post-processing ($t=1.0$), SynthCP achieves higher AUPR, but also produces more false positives, resulting in degrading FPR95 and AUROC. After pruning out false positives at high MSP positions ($p^{(i)}>0.999$), we achieved the state-of-the-art performances under all three metrics.

\begin{table}[h]
\caption{Performance change by varying post-processing threshold $t$} \label{tab:th_post_process}
    \centering
    \begin{tabular}{p{0.15\textwidth} | p{0.1\textwidth}|p{0.1\textwidth} |p{0.1\textwidth} |p{0.1\textwidth} |p{0.1\textwidth}}
        \hline
        t & 0.8 & 0.9 & 0.99 & 0.999 & 1.0\\
        \hline
        FPR95 $\downarrow$   & \textbf{28.6}  & 28.5 & 28.2 & 28.4 & 46.0\\
        AUROC $\uparrow$     & 88.3  & 88.4 & \textbf{88.6} & 88.5 & 81.9\\
        AUPR $\uparrow$       & 7.4   & 7.7 & 8.8  & \textbf{9.3} & 8.1\\
        \hline
        \end{tabular}
    \label{tab:t}
\end{table}


\section{Conclusions}

We present a unified framework, SynthCP, to detect failures and anomalies for semantic segmentation, which consists of an image synthesize module and a comparison module. We model the image synthesize module with a semantic-to-image conditional GAN (cGAN) and train it on label-image pairs. We then use it to reconstruct the image based on the predicted segmentation mask. The synthesized image and the original image are fed forward to the comparison module and output either failure detection (both image-level and pixel-level) or the mask of anomalous objects, depending on the specific task. SynthCP achieved the state-of-the-art performances on three challenging datasets.

\noindent\textbf{Acknowledgement} This work was supported by NSF BCS-1827427, the Lustgarten Foundation for Pancreatic Cancer Research and NSFC No. 61672336. We also thank the constructive suggestions from Dr. Chenxi Liu, Qing Liu and Huiyu Wang.


\clearpage
%
%
\bibliographystyle{splncs04}
\bibliography{egbib}
\end{document}